# Right whale recognition using convolutional neural networks

Andrei Polzounov, Ilmira Terpugova, Deividas Skiparis, Andrei Mihai

{andrei.polzounov, ilmira.terpugova, deividas.skiparis, andrei.mihai}@est.fib.upc.edu

Facultat d'Informàtica de Barcelona, Universitat Politècnica de Catalunya

## Abstract

We studied the feasibility of recognizing individual right whales (Eubalaena glacialis) using convolutional neural networks. Prior studies have shown that CNNs can be used in wide range of classification and categorization tasks such as automated human face recognition. To test applicability of deep learning to whale recognition we have developed several models based on best practices from literature. Here, we describe the performance of the models. We conclude that machine recognition of whales is feasible and comment on the difficulty of the problem.

## 1. Introduction

There are fewer than 500 North Atlantic right whales (Eubalaena glacialis) remaining in the world (Fujiwara and Caswell [9]). The species is highly endangered and is considered as such by the U.S. and Canadian governments. Recognizing individual whale specimens is important if we are to help the species recover to sustainability. Recognizing individual specimens from shipborne or helicopter imagery is a tedious task for marine biologists. Convolutional neural networks are quickly becoming the tool of choice for automated image recognition and classification [1, 2, 3, 4, 5, 6].To the best of our knowledge automated "face" recognition techniques have not previously been proposed for recognizing right whales. In this report we summarize our analysis of using convolutional neural networks to recognize individual North Atlantic right whales. The dataset has been put together by NOAA Fisheries (National Oceanic and Atmospheric Administration) [8] and distributed as a Kaggle challenge.

### Convolutional Neural Networks

In the last few years deep convolutional neural networks have been seeing an explosion in literature and on the internet. They differ from traditional networks by making the explicit assumption that the input data is an image. This allows convolutional networks draw another inspiration from nature – receptive fields of vision. CNNs can focus on a specific parts of the image using convolutions.

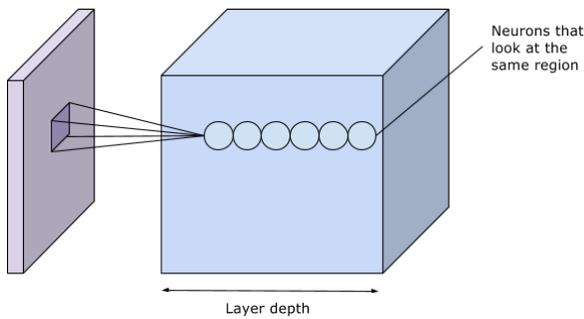 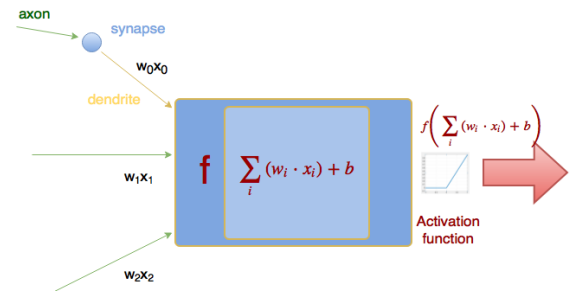

*Figure 1: Receptive field of neurons in CNN*          *Figure 2: Traditional neuron activation mechanism*

CNNs consist of convolutional layers – which act as receptive fields, followed by pooling layers – which decrease the amount of features and pixels the next convolutional and pooling layers can focus on. Convolutional and pooling layers are stacked many times until finally connected to a classical neural network (named fully connected in deep learning literature) with some hidden layers and then finally the output layer as the classifier.

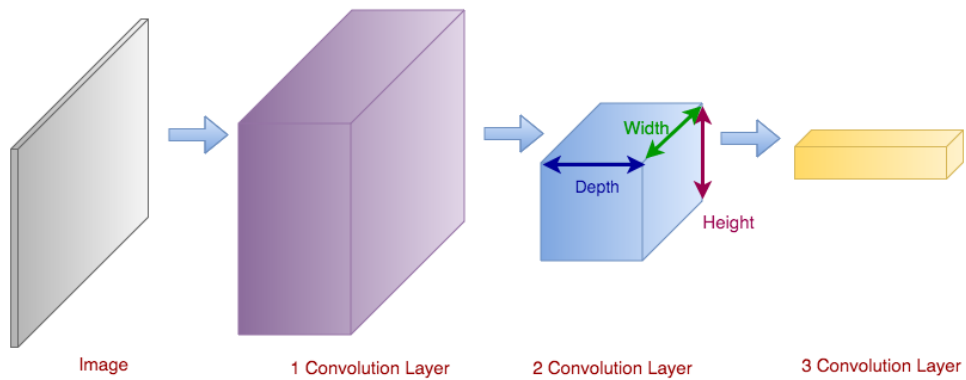

*Figure 3: CNNs arrange the neurons in 3 Dimensions (width, height and depth)*

Theoretically recognizable features should be close to each other – this allows to create CNN receptive fields using fully connected neural networks in which the first neuron layer is connected to every single pixel of the image. This will result in a potentially higher descriptive ability of the network at the cost greatly increased complexity. The downside of fully connected NNs is that they do not exploit locality of related features as CNNs do (features far away from each other are not likely to be related). See Appendix A for more CNN theory.

## Successful Applications of Conv Nets

Deep networks have been very successful in classifying many different kinds of objects. Some of the best networks are able to classify as many as 22000 different categories learned from a set of 15 million images (Krizhevsky et al [1] and Russakovsky et al [6]). Despite highly optimized code and 3 high performance video cards – the ImageNet network takes about 5 days to train according to the authors [1].

Another very successful network is outlined in a paper from Google – the GoogLeNet [2]. It employs many classical computer vision techniques along with the raw computational power of CNN (convolutional neural networks). Some of the novel techniques utilized in the architecture from Google are using stacks of very small convolutional kernels instead of larger kernels. Smaller kernels are easier to train individually and when stacked they can provide the same discrimination of features as larger filter. These small kernels are further popularized by the work of Simonyan and Zisserman [4] and are in fact what the winners of this Kaggle challenge used [18] to classify the right whales.

## 2. Preprocessing of imagery

*Expected features for detecting individual whales*

The feature that we are most interested is called the callosity pattern (fig. 4) which includes the facial markings on top of the head of the whale and the white markings above the blowholes. These features are unique to each whale. Some of the features that we decided to ignore were the shape of the tail, dorsal fins and side flippers which may also have been useful, but would have increased the complexity of the detection.

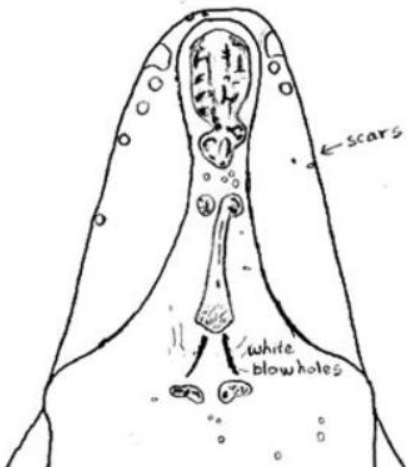
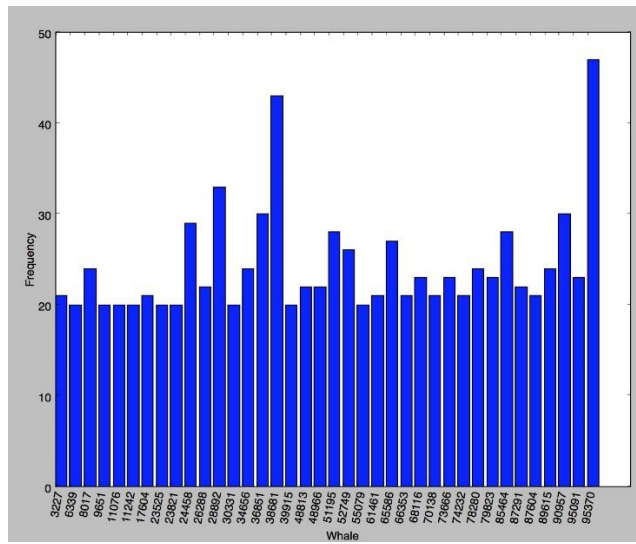

*Figure 4: Unique callosity pattern of an individual whale specimen (NOAA Fisheries Whale Catalog, 1997 [8])*

*Figure 5: Histogram of label distribution in α-whales data*

## The dataset

From the original dataset (NOAA Fisheries [8]) of 11469 images only 4542 images were labeled; the labels included 427 unique individuals. Some of the labels only included a single image of the whale. To limit training time we extracted a new dataset – α-whales from the labeled data (fig. 5). Taking only specimens which have 20 or more labeled images, this gave us a set of 924 images of 38 unique whales, which is what we based our classifier on.

*Dataset Preprocessing*

The raw images from the dataset were a very large resolution. Operating on such images would require massive processing power. In addition much of the visible area of each image was taken up by the water. A large amount of noise with respect to the ROI (region of interest) was added by the waves and splashes around the whale.

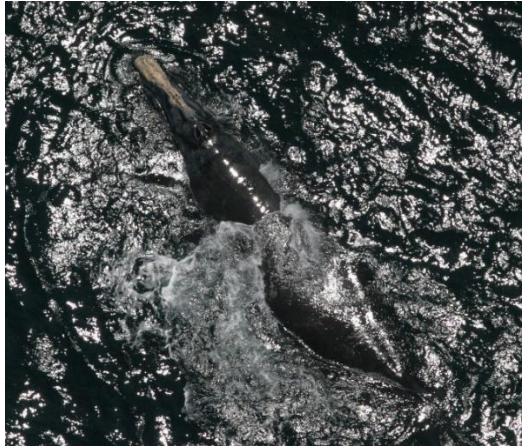

*Figure 6: Example of noise from waves and splashes*

To preprocess the data we had tried to segment the ROI of the whale from the water. We managed to discard the majority of the water pixels by segmenting on the histogram of the saturation channel of the image.

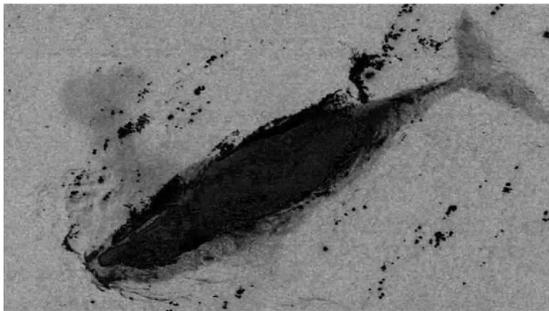
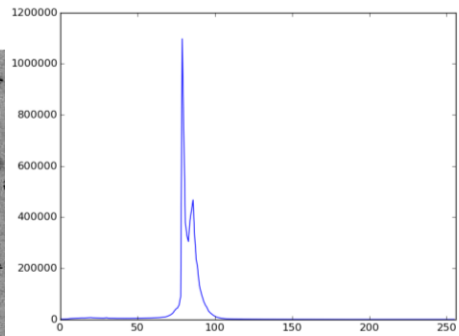

*Figure 7: Saturation from HSV channel*           *Figure 8: Histogram of saturation channel*

Ideally the saturation histogram should have two visible peaks (as above). The first and greater maxima is a marker for the water pixels which make up the majority of the image. The second and lesser maxima is therefore expected to be the whale or the foreground of the image. We can threshold the image using the minima that can be found between the two local maxima points – which leaves us with the pixels corresponding to the whale and the surrounding noisy pixels of waves/splashes.

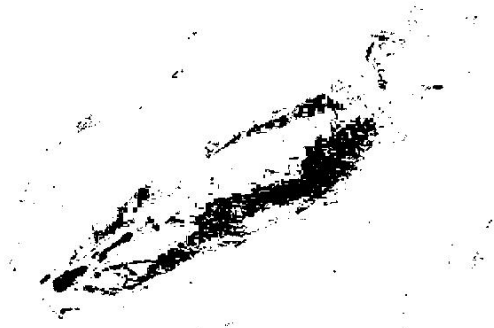

*Figure 9: Extracted mask of ROI from thresholding the saturation channel*

Using this ROI polygon we can fix yet another problem with the data – whales all face random directions. A solution to this particular problem is to inscribe an ellipse into the ROI polygon. The major axis of this polygon will roughly coincide with the major axis (head to tail) of the whale. Knowing the angle of rotation we can use an affine transformation to rotate the image to our desired location. The whale will now be facing either up or down.

$$\begin{bmatrix} x' \\ y' \end{bmatrix} = \begin{bmatrix} cos\theta & -sin\theta \\ sin\theta & cos\theta \end{bmatrix} \begin{bmatrix} x \\ y \end{bmatrix}$$

The entire process works well about 80% of the time, and even in successful runs of the preprocessor we could not know whether the whale was facing up or down. Since we were not able to reliably preprocess the raw images using conventional CV (computer vision) techniques we turned to manual preprocessing:
- Rotate the image to have the whale facing right.
- Cut out a square area with the whale dead center.
- Resize image to 256x256

These steps ensured that we obtained the "passport" photos of that the CV preprocessor should have made.

## 3. Experiments with CNN models

As part of the project we have evaluated many different CNN models. The biggest constraint of trying new models and tuning the associated hyper-parameters were training time, processing power and lack of sufficient computer memory. Due to design of Google TensorFlow a large chunk of memory is pre-allocated to be used in its computation graph [20]. None of our models could fit into 2 GB GPU memory that was available, which could have provided a huge speedup over training on CPU.

Model topology diagrams can be found in Appendix B.

1) <u>DumbNet</u> – our first model (retroactively named DumbNet) was completely of our own design. It consisted of 5 convolutions each followed by a pooling layer. The model heavily relied on pooling (once after every convolution) and many of the neurons died out from over-saturation. A simplified version of the network managed to minimize cross entropy loss and achieved 7.3% validation error. We concluded that the model was not complex enough as it failed to converge during training.
2) <u>AlexNet</u> (Krizhevsky et al [1]) – this network was difficult to work with due to the non-standard convolution and pooling layers which changed the size of the output image in a way very different from other networks. This network did not converge either.
3) <u>VGGNet</u> (Simonyan and Zisserman [4]) – operates on the use of small stacked convolutions with fewer pooling layers in between. The authors argue that a stack of three 3x3 convolutions activated by ReLU (rectified linear unit) [11] activations can be more discriminative than a single 7x7 convolution. We experienced problems running VGG because of the massive computational cost. It consists of 13 convolutional layers, 5 pooling layers and 2 wide fully connected layers before finally coming to the classifier neuron layer. Our computers were not able to reasonably run the network.

4) <u>DeepSenseNet</u> – this network was inspired by the winner of this Kaggle competition [19]. The authors must have themselves drawn inspiration from VGG as the network seems to be a simpler version of that one. We were able to obtain adequate results from our interpretation of this topology after about 12 hours of training. We used an exponentially decaying learning rate and local-response normalizations [1] of the activation levels of every convolutional layer. The results were over-fit to the training data with about 80% classification accuracy on the training set and 15% on the validation set. This is enough to prove statistical significance of the classifier.

## 4. Results

To set the baseline for neural networks, k – nearest neighbor (kNN) [12] classifier with cropped images has been used. The cropped and rotated images containing only the ROI of the whale were used (image size 256 x 256. See dataset preprocessing section for details).

kNN classification was applied to raw feature vectors i.e. vector of unrolled image pixel values. A number of different k values have been used, namely k=1, 3 and 5. Euclidean distance was used as a measure of similarity. The same train and validation data split as for CNN has been used with the kNN classifier to ensure that the results would be comparable.
The best accuracy achieved was 0.2097 ($\sigma = 0$ from 20 runs) for k=1. Full table below.

In order to improve accuracy of the naive kNN classifier we have also implemented PCA (principal component analysis) (Jolliffe [13]) and LDA (linear discriminant analysis) algorithms [14, 15] to reduce feature size before comparison. Lastly we changed the distance metric from Euclidean to Chebyshev distance to improve kNN results. See Appendix A for in depth analysis of PCA, LDA and kNN.

Below is the full table of kNN classification results:

| k | RAW | PCA | PCA+LDA | PCA+LDA (Chebyshev) |
|---|---|---|---|---|
| k=1 | 0.2097 | 0.2419 | 0.5726 | 0.5968 |
| k=3 | 0.1290 | 0.1371 | 0.5484 | 0.6048 |
| k=5 | 0.1290 | 0.1452 | 0.5242 | 0.6129 |

*Figure 10: kNN classification results. #Runs = 20 per k*

The result of training and evaluating ConvNets can be found below in figure 11. Some of the networks did not converge or could not be evaluated and are noted as such in the table of results.

| Network | Learning rate (starting) | Dropout | Batch size | Norm | Min cross-entropy | Validation accuracy |
|---|---|---|---|---|---|---|
| *DumbNet* | $10^{-3}$ | 0 | 20 | LRN | 621.7 | Insignificant |
| | $10^{-4}$ | 0 | 20 | LRN | 621.7 | Insignificant |
| | $10^{-5}$ | 0 | 20 | LRN | 617.6 | Insignificant |
| | $10^{-4}$ | 0 | 10 | L2 | 214.3 | Insignificant |
| | $10^{-4}$ | 0 | 10 | LRN | 310.8 | Insignificant |
| *DumbNetSimple* | $10^{-4}$ | 0 | 10 | LRN | 28.5 | 7.3% |
| *AlexNet* | $10^{-4}$ | 0 | 10 | LRN | 230.3 | Insignificant |
| | $10^{-5}$ | 0 | 10 | LRN | 207.2 | Insignificant |
| *VggNet* | $10^{-4}$ | 0.5 | 10 | - | - | Out of memory |
| *VggLikeNet* | $10^{-4}$ | 0 | 20 | LRN | 656.2 | Insignificant |
| *DeepSenseNet* | $10^{-4}$ | 0 | 20 | L2 | 646.4 | Insignificant |
| | $10^{-4}$ | 0.5 | 20 | LRN | 274.6 | Insignificant |
| | $10^{-4}$ | 0.5 | 20 | L2 | 557.4 | 3.3% |
| | $10^{-4}$ | 0 | 20 | LRN | 138.9 | 15.3% |

*Figure 11: CNN classification results*

## 5. Analysis

*Successes*

We can demonstrate that for simple problems such as the MNIST handwritten digit recognition (Le Cun et al [5]) (see Appendix D for our results applied to the MNIST problem), we can easily outperform the naive kNN approach using a CNN. Even though we have failed to demonstrate better performance classifying our α-whales dataset using a CNN vs the kNN we can still produce statistically significant results of about 0.15 classification accuracy on the validation set. This is much greater than a random guess of $1/38 = 0.0263$, but not as high as the 0.2 value from kNN.

We have been able to show that many of the filters that we trained with our convolutional layers are in fact detecting the callosity patterns of the whales with different levels of activations (close to zero or closer to one) and with different types of filters. Some of the filters act as high pass filters – which act edge detectors and some as low pass smoothing filters.

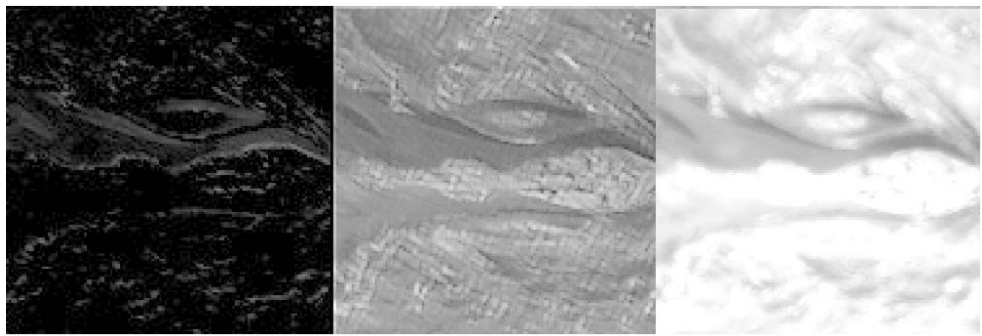

*Figure 12: Output of a sample neuron layer.*

(a) High pass filter. Low activation levels overall (close to zero mean) does not show the callosity pattern at all and the activated patterns are mostly noise from the waves. Perhaps the outline of the whale could be useful to deeper layers.
(b) Low pass filter. Average activation levels (as seen by the gray levels), very clear callosity pattern and the features from the waves have been smoothed.
(c) Low pass filter. Very high overall activation. Similar features to (b) with a good callosity pattern but with slightly more noise – this filter is probably still useful.

Another success of the project is to note that we have qualitatively verified that stacks of small convolutions such as the ones proposed by (Szegedy et al [2], Simonyan and Zisserman [4]) are easier to work with and converge much faster than networks with large convolution kernels. We suspect that many successful projects in the short term will adopt a similar approach, in fact the winner of the Kaggle competition – Deep Sense used a simplified VGG net [19].

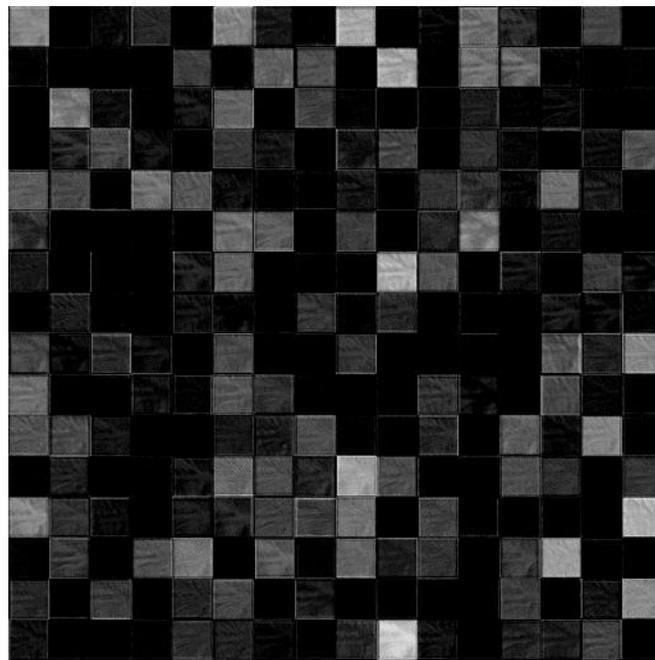

*Figure 13: Filtered images shown as output of a convolutional layer. See C.3 In Appendix C for larger example*

## Shortcomings

Our DeepSenseNet was overfit to the training data, this can be easily interpreted from the high classification accuracy on the training set and low accuracy on validation data. Another major problem is the arbitrary validation set, we did not use k fold cross-validation as it would have required multiplying the already long training time by a factor of k. Furthermore, we have observed saturation in some of the convolutional layers even after normalizing each one of them.

Each one of the squares in figure 13 corresponds to a feature detected by a feature in the conv net. Many of the squares are completely black meaning the neurons are dead and cannot be brought back even with very strong multipliers. This is because of a known property of the ReLU activation function of multiplying negative result by zero (Nair and Hinton [11]). Other activation functions have other downsides that were also considered during the design phase of this project. For further discussion of training convolutional filters see Visualizing Conv Nets in Appendix C.

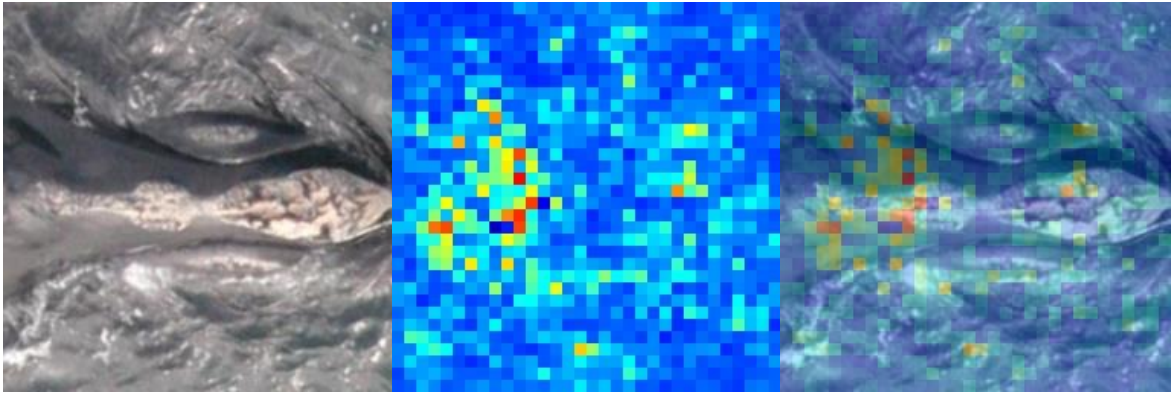
*Figure 14: Saliency map of a whale displaying the most important features*

To analyze the features that our network was learning from, we drew a saliency map of a sample image (fig. 14) as described by Zeiler and Fergus [21]. The map can be interpreted as a heat map with cold (blue) pixels contributing less to feature classification than the hot (red) pixels. The results were disappointing as the hot areas of the saliency maps were showing that noisy areas were contributing more to classification than the desired features. Ideally the saliency heat map should only display the callosity pattern of the whale. For a further discussion on saliency maps see Appendix C.

## Possible Improvements
### Preprocessing
We were unable to perfect our preprocessing algorithms using conventional computer vision techniques. A solution to utilize computational intelligence to the whale detector. We could take the full set of 11000 images and manually tag the location of the whales head on some of them. Using this supervised data a "whale-head-detector" can be trained. This algorithm combined with our working rotation algorithm will result in a whale that is roughly aligned to the horizontal axis. At this point the whale will be facing either left or right and another classifier can be trained to flip the images to their correct orientation.

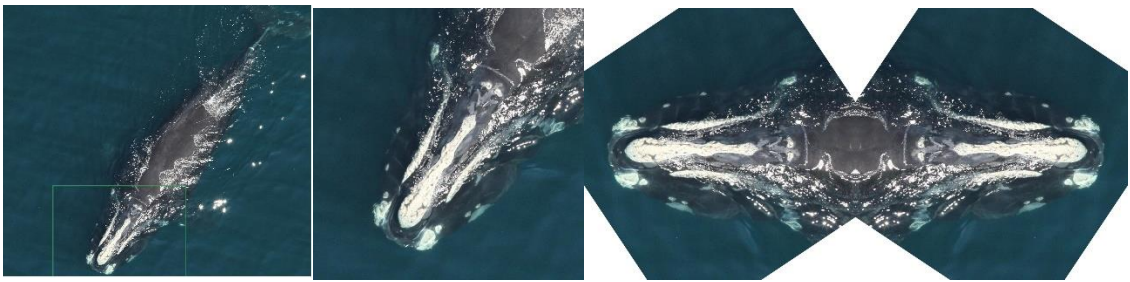
*Figure 15: Proposed whale image preprocessing system*

This working preprocessor can then be run on the full set of data – labeled and unlabeled. Ideally this larger dataset should be more trainable without overfitting the data then our α-whales dataset.

### Augmenting the training data

After obtaining the full training data we would be left with 4500 labeled images of 427 whales – still not enough data to detect some of the whales. We can expand the labeled images set using some augmentation techniques. For every image we would run it through some filters which are meant to signify the variance between the images of the set and then use filter outputs as additional training data. Some of the filters we would use: low pass filter (smoothing), high pass filter (edge detection or sharpening) and various affine transforms such as rotation about the horizontal axis, scaling and pixel-wise shift.

### Other promising techniques

It has been shown that a Support Vector Machine (SVM) can often perform better than a fully connected neural network. The SVM can potentially replace the fully connected layer that follows convolutional layers (Huang and Le Cun [22]). Another idea is to perform ZCA (zero component analysis) on the imagery as preprocessing. The ZCA transform should minimize cross-correlation of features.

### Improved hardware

A huge weakness of our project was the lack of cross validation and the constraints imposed on us not being able to train more complex networks in a reasonable amount of time. Training on a video card with a large amount of memory or on the cloud would enable us to iterate through hypothesis much faster.

## 6. Conclusion

We have confirmed that CNNs classify images that pose significant difficulty to untrained humans. While the results were less than ideal they were sufficient to show the feasibility of the solution. We conclude that the problem is too computationally intensive to solve using unprocessed imagery. Therefore we propose a way of chaining machine learning driven preprocessing system. We have compared many different neural topologies and we concur with the trend in literature of using small kerneled stacked convolutions. Ultimately we must note that the heavy computational requirements remain a limiting factor for the usefulness of deep learning techniques in image recognition.

# Appendix A: Theoretical Concepts

*Baseline Algorithm – kNN (k Nearest Neighbors)*
In order to compare our neural network performance, we have built a kNN classifier. This classifier builds a model from training images by arranging them in a d-dimensional space by some distance measure. In order to classify a test image, it then finds k training images, which are closest to a test image and uses most frequent label as the solution [12].

Each image in the set is represented by a feature vector (FV). FV in the simplest form could be a vector of unrolled grayscale image pixel values. For an image with dimensions $[h\ w]$, the FV is $[1\ w*h]$.

The performance of kNN classifier on raw data is likely to be very poor. In order to achieve better classification accuracy of the kNN classifier, two different dimensionality reduction and feature extraction techniques can be used:
1. Principal Component Analysis (PCA) - Aims to explain variance in data by transforming the data to a new set of uncorrelated features, the principal components (PCs) [13]. PCs have much lower dimensionality and preserve all of the original variance. They are ranked (descending order) by the amount of variance they explain, so removing last n PCs, most likely noise and unimportant data is scrapped.
2. Linear Discriminant Analysis (LDA) – a feature extraction technique originally developed by R. A. Fisher in 1936. The algorithm is based on searching for a linear combination of variables that best separate two classes. A generalized version of LDA [14] has been used, which allows working with multiple classes. LDA is similar to PCA in the sense, that they both try to extract linear combinations of variables which best explain the data [15]. However LDA attempts to model differences between different classes of data, which seems to be a perfect tool for trying to distinguish between faces of different whales.

*Backpropagation*
The backpropagation algorithm is used while training to propagate the error corrections from the output layer down through all the connected layers of neurons. This will ensure that the output result on the next epoch of training will provide a better score.

The main idea of the algorithm is using gradient descent to calculate the required changes of the current weights to reach a value with a minimal error difference from the required output. If the output vector was defined to be a dot product of the weights vector and the input vector the following equation would be found:
$$\boldsymbol{o(x) = w \cdot x}$$

The training error of this linear unit (preceptor without the thresholding function) can be calculated by many functions – a simple way is summing the squared difference (t – refers to ideal output):
$$\boldsymbol{E(w) = \frac{1}{2}\sum(t-o)^2}$$

It is important to realize the correspondence of the weights to the training error will be some sort of hyperparabola – the specific shape of which will of course depend on the dataset. Here is an image of one such imagined parabola with only two weights. The global minima of the parabola will correspond to the best possible weights.

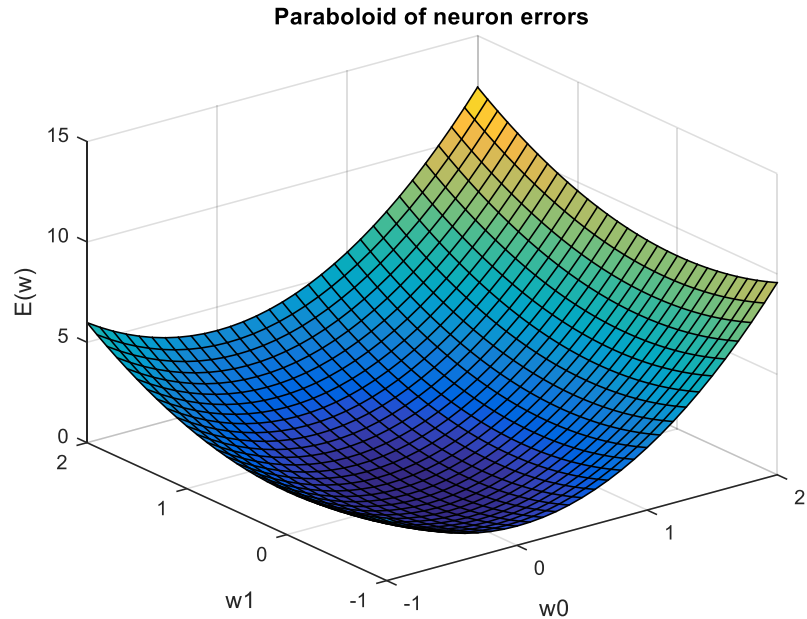

*Figure 14: Possible paraboloid graph of neuron errors w.r.t. the weight values*

Taking the gradient of the error vector with respect to each of the weights:

$$\nabla E(w) = \left[\frac{\partial E}{\partial w_0}, \frac{\partial E}{\partial w_1}, \dots, \frac{\partial E}{\partial w_n}\right]$$

If one were to evaluate $\nabla E(w)$ with the current weights – the result would be a vector showing the steepest increase of slope away from the current point. The negative of that would be the steepest decrease. Therefore new weights can be updated with the following delta:

$$\Delta w_t = -\eta \nabla E(w_{t-1})$$

$\eta$ is the learning rate. Setting $\eta$ to a greater value can result in a quicker convergence, however it can overshoot the global minima, while setting it too low will result in a very slow training process.

Gradient descent is not guaranteed to reach the global minima, however this is less of a problem when N is large for N-dimensional data as there will be more "routes" of sliding down towards the minima with a larger number of weights. Backpropagation is simply applying gradient descent recursively from the outermost layer all the way through the network.

*Adam Optimizer*

An improvement on the classical gradient descent algorithm would be to maintain inertia while descending towards the minima. The simple gradient descent algorithm can be prone to radically changing directions of descent if the data is noisy. 1st and 2nd order moments can be added to increase the chance of staying on the true path while training.

The Adam (adaptive moment estimation) algorithm is described by (Kingma & Ba, 2014). It is a state of the art method for training. It utilizes two extra variables while calculating the weight updates:

$m_t$ – exponential moving average of the gradient ($\nabla E(w)$)
$v_t$ – exponential moving average of the elementwise squared gradient ($(\nabla E(w))^2$)

On the first iteration both of the vectors are set to zeroes. After wards they will be calculated using their corresponding exponential decay rates $\beta_1$ and $\beta_2$ and epsilon $\epsilon$.

On each iteration of the Adam gradient descent first the biased moment estimates are found:

$$m_t = \beta_1 m_{t-1} + (1 - \beta_1)\nabla E(w)$$
$$v_t = \beta_2 v_{t-1} + (1 - \beta_2)(\nabla E(w))^2$$

The next step is to compute bias corrected moment estimates by dividing by one minus β to the power of t:

$$\widehat{m_t} = \frac{m_t}{(1 - \beta_1^t)}$$
$$\widehat{v_t} = \frac{v_t}{(1 - \beta_2^t)}$$

Finally the new weights can be recalculated:

$$\boldsymbol{w}_t = \boldsymbol{w}_{t-1} - \frac{\eta \widehat{m_t}}{(\sqrt{\widehat{v_t}} + \epsilon)}$$

*Soft-max activation for output neurons*

The output layer of neurons is used for classification, however all of the neurons may have different activation levels. In order to estimate the activations as probabilities ($P \epsilon [0,1]$) we need to make sure that all of the outputs sum to one. This can be achieved with the soft-max activation function:

$$g(h_k) = \frac{\exp(h_k)}{\sum_{k=1}^{N} \exp(h_k)}$$

*Regularization*

In order to prevent overfitting we used dropout of neurons with 0.5 probability of dropout in the fully connected layers. The activation levels of the convolutional layers were regularized with L2 normalization or with local response normalization outlined by Krizhevsky et al [1].

## Appendix B: Conv Net Topologies

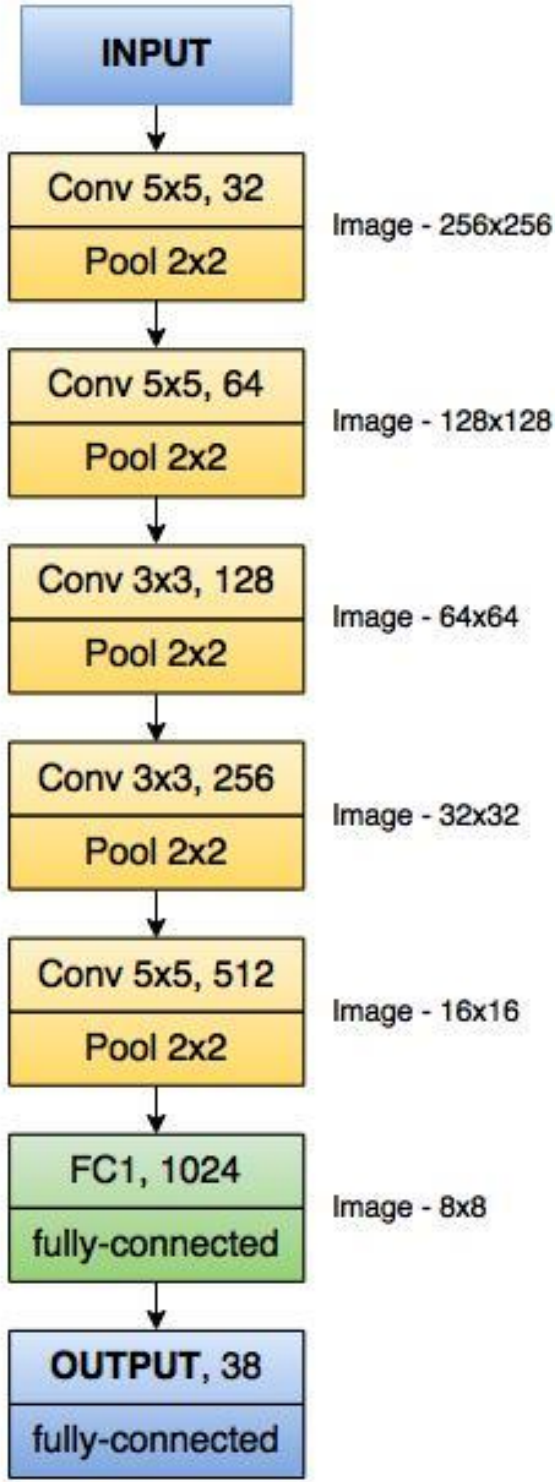

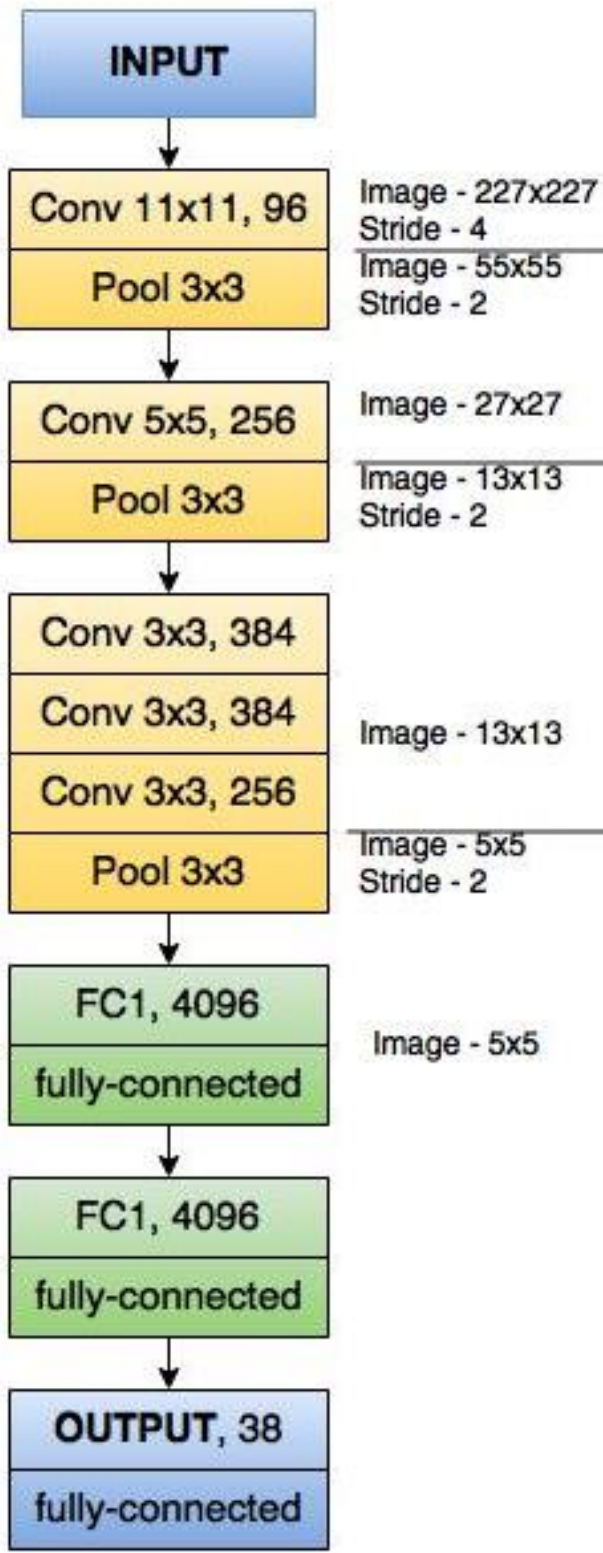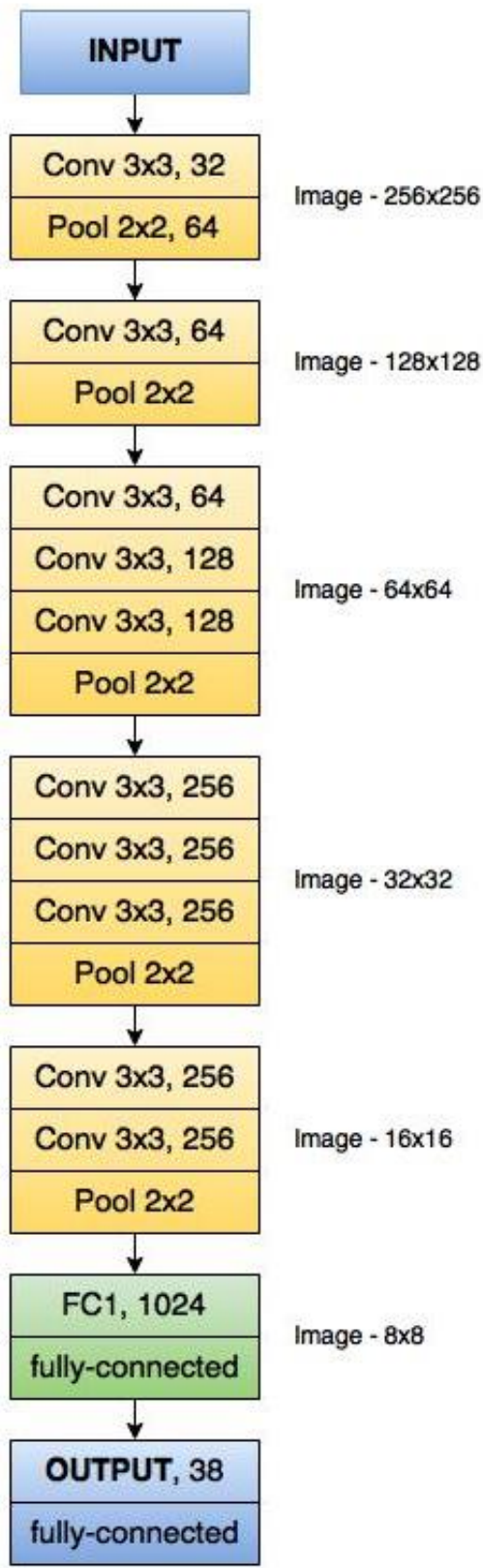

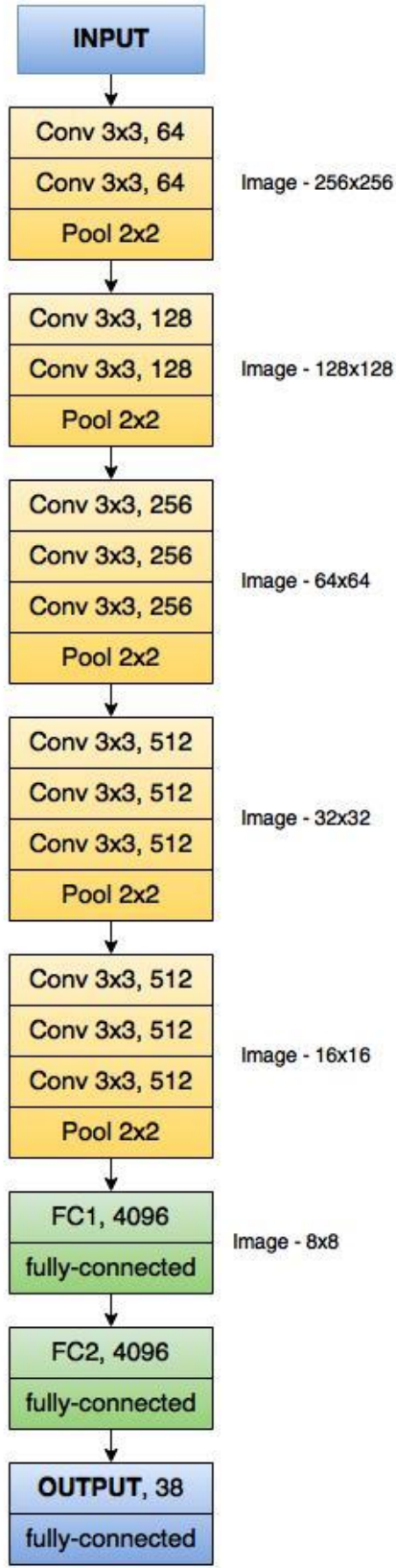 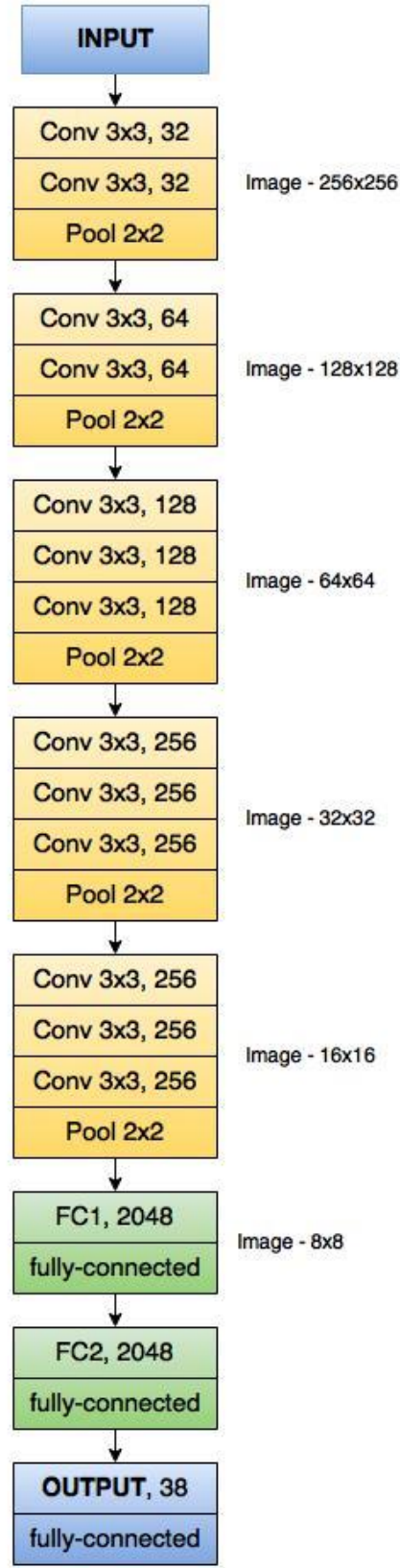

# Appendix C: Visualizing Conv Nets
*Saliency Map*

When a deep convolutional network is trained for classification it should learn to pick up the parts that are distinguishable between classes. In our case we expect that the network will get the information from callosity patterns of whales. To check what parts of images are used in classifier we can do the following:

1. In the current image compute the probability of correct class.
2. Occlude small part of the image, by setting its intensity to 0 (fig.1).
3. Get the output for occluded image from classifier.
4. Compute the difference between the probabilities for the whole image and occluded one. This difference will be the measure of importance of occluded part for classification. The bigger is the difference the bigger is contribution of the region to the final score.
5. Repeat the same procedure for next square box.

The 'heat map' of different images is shown on figure 2. The red pixels correspond to most important parts.

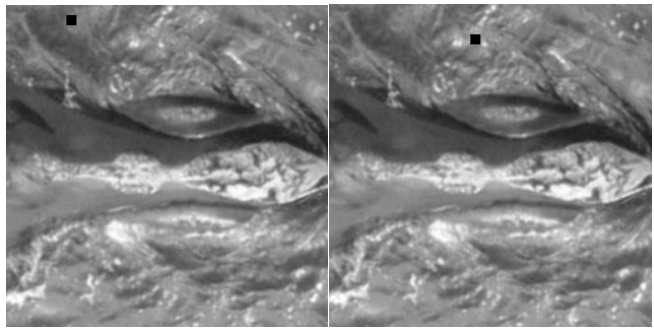

*Figure C.15: Occlusion in input image*

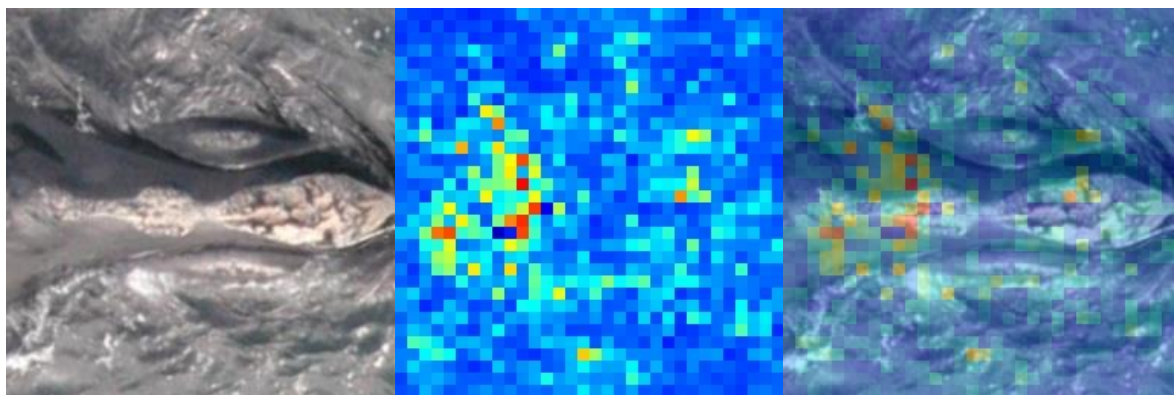

*Figure C.16: Saliency Map*

We can see from salience map that the network is not trained well to distinguish only the callosity pattern. It also takes into account some noise from water and splashes.

*Visualization of filters*

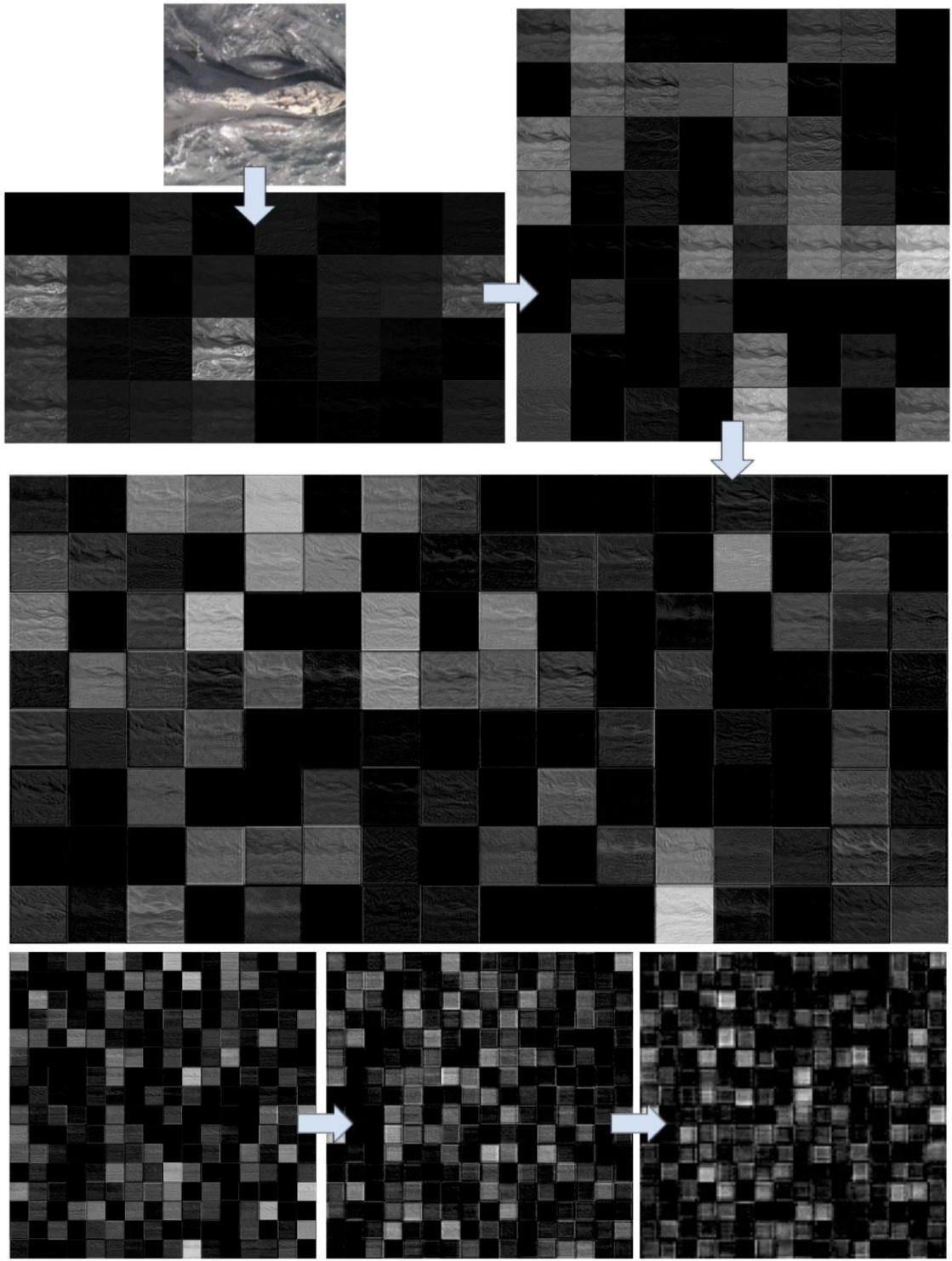

*Figure C.17: Output of convolutional layers*

As long as the networks use shared parameters for each depth layer of neurons the output of each layer will consist of n filtered images, where n is the depth of the layer. We use these filtered images to understand how the network is actually processing the input. The example of an activation output for each layer is shown in figure 3.

We can see that some of the more successful neurons filter out almost all the water noise and leave only the whale callosity pattern.

There is another use of the neuron visualizations. The networks use ReLU activation function which is simply thresholding output at zero. It was found that this type of activation accelerates the convergence and provides much faster computations comparing to sigmoid functions. But it has a problem of 'dying neurons' when some of the weights could be updated in such a way that the neuron will never activate with any input again.

The images of output can help to find out these 'dead' neurons. If we compare the outputs of the neurons for different input images we can discover which neurons don't activate (figure 4).

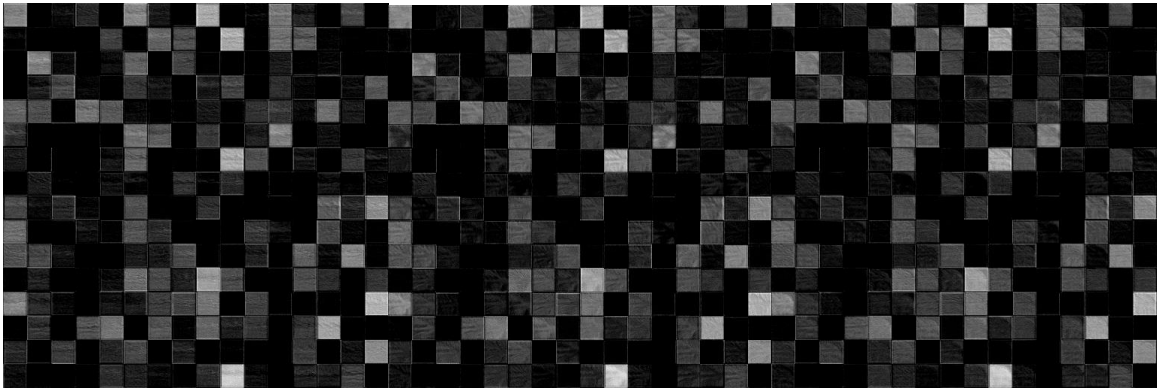

*Figure C.18: Output from layer 4 for different images*

## Appendix D: MNIST Mini Problem

A simple way of illustrating how CNNs perform compared to more naïve methodologies is to show how CNNs work on a simpler problem. The MNIST handwritten digit dataset was chosen for this problem. The dataset consists of 60 thousand training images of 10 classes and 10 thousand of test images. Individual image size is 28x28 pixels.

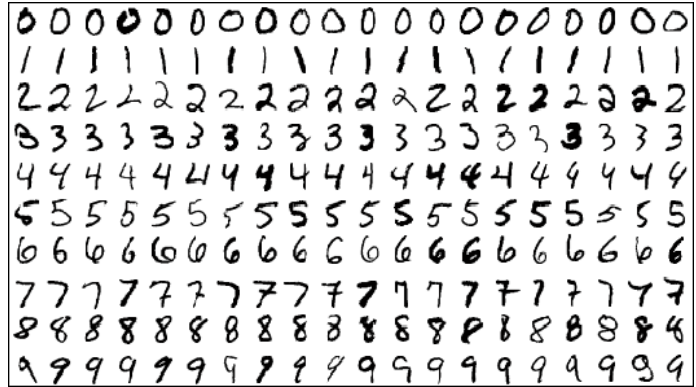

*Figure D.1: Some MNIST digit images*

A network with two convolutional layers was used, each of which was followed by a pooling layer, and one fully connected layer (figure 2). Neuron dropout was applied in the fully connected layer to avoid overfitting.

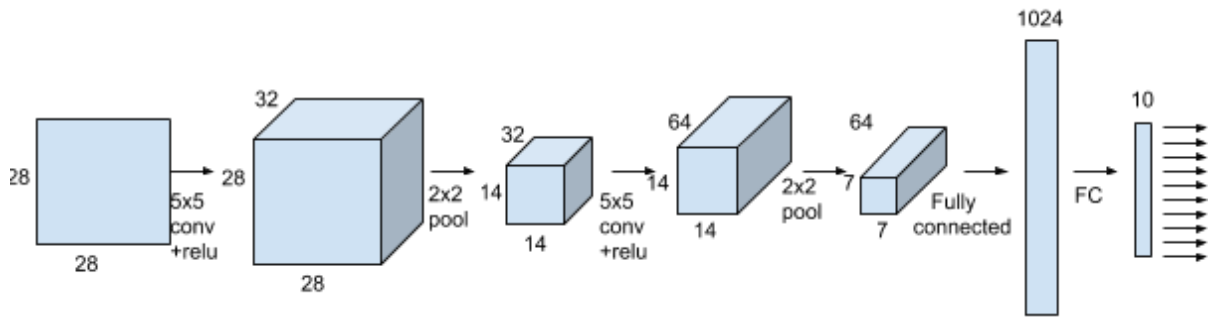

*Figure D.2: Network architecture for MNIST classifier CNN*

A k nearest neighbor classifier is used to compare the performance of the CNN. The k-NN classifier uses the following parameters: k = 5 and Minkowski distance.
It is compared to the neural network (fig. 2) trained with 20000 steps, mini batch size 50, learning rate $10^{-4}$ and dropout probability 0.5. Using 10-fold cross validation on the data set we were able to calculate and compare accuracy of kNN vs the CNN implementation.

The accuracy for k-NN was 96.88%. With the CNN we obtain an accuracy of 99.18% which is approaching human error rate.